\documentclass[letterpaper]{article} 
\usepackage{aaai24}  
\usepackage{times}  
\usepackage{helvet}  
\usepackage{courier}  
\usepackage[hyphens]{url}  
\usepackage{graphicx} 
\usepackage{natbib}  
\usepackage{caption} 
\usepackage{soul}
\usepackage{url}
\usepackage{graphicx}
\usepackage{amsmath}
\usepackage{booktabs}
\usepackage{algorithm}
\usepackage{algorithmic}
\usepackage{tikz} 
\usepackage{amssymb}
\usepackage{xspace}
\usepackage{xcolor}
\usepackage{pifont}
\usepackage{algorithm}
\usepackage{subfigure}
\usepackage{amsmath,amsfonts}
\usepackage{algorithmic}
\usepackage{enumitem}

\newtheorem{lemma}{Lemma}

\newcommand{\ourmethod}{{\color{black}\ensuremath{\texttt{\textsc{Coin}}}}}

\newcommand{\greencheck}{{\checkmark}}
\newcommand{\xmark}{\ding{55}}%
\newcommand{\redx}{{\xmark}}

\newcommand{\eg}{\emph{e.g.}\xspace}
\newcommand{\ie}{\emph{i.e.}\xspace}

\title{\ourmethod: Chance-Constrained Imitation Learning for Uncertainty-aware\\ Adaptive Resource Oversubscription Policy}
\author{Lu Wang$^*$, Mayukh Das$^\mathsection$, Fangkai Yang$^*$,  Chao Duo$^*$, Bo Qiao$^*$, Hang Dong$^*$, Si Qin$^*$, \\Chetan Bansal$^\mathsection$,  Qingwei Lin$^*$, Saravan Rajmohan$^\mathsection$, Dongmei Zhang$^*$, Qi Zhang$^*$}
\affiliations{
     Microsoft$^*$, Microsoft 365$^\mathsection$
}

\begin{document}
\maketitle
 \begin{abstract} 
  We address the challenge of learning safe and robust decision policies in presence of uncertainty in context of the real scientific problem of adaptive resource oversubscription to enhance resource efficiency while ensuring safety against resource congestion risk.
  Traditional supervised prediction or forecasting models are ineffective in learning adaptive policies whereas standard online optimization or reinforcement learning is difficult to deploy on real systems. Offline methods such as imitation learning (IL) are ideal since we can directly leverage historical resource usage telemetry. But, the underlying aleatoric uncertainty in such telemetry is a critical bottleneck.
  We solve this with our proposed novel chance-constrained imitation learning framework, which ensures implicit safety against uncertainty in a principled manner via a combination of stochastic (chance) constraints on resource congestion risk and ensemble value functions. This leads to substantial ($\approx 3-4\times$) improvement in resource efficiency and safety in many oversubscription scenarios, including resource management in cloud services.
\end{abstract}


\section{Introduction}

Resource \textit{''Oversubscription"} is widely practised across the professional services and service products industry, including cloud services, airlines, logistics etc., to optimize their operating costs (capacity efficiency) via `thin-provisioning' while ensuring service reliability.
The key idea is that a system offers more resources to users or entities than its available capacity, assuming not all users would simultaneously or fully utilize their allocated capacity in order to diminish the sum of unutilized
resources and thus increase profits. 
For example, cloud providers oversubscribe their virtual computing resources to minimize stranded resources in order to maximize profits~\cite{baset2012towards,breitgand2012improving} [details in Appendix A.1]. 
In the airline industry, 
overbooking (oversubscribing) fight tickets beyond aircraft capacity is quite common and helps improve airline load factors and reduce revenue losses due to cancellations and no-shows~\cite{airlineoverbooking1975,nazifi2021proactive}. 

Oversubscription policy design (whether, when or how much resource margin to oversubscribe for different sets of users) is a critical (often overlooked) scientific problem and is structurally/semantically different from resource allocation problem.
Conservative policies lead to inefficient resource  utilization and wastage, whereas aggressive policies cause resource congestion, compromising service reliability. Therefore, \ul{oversubscription policies should be \textbf{adaptive} to the dynamic patterns of resources usage} which is a sequential decision-making problem. However, adaptive policies can be unsafe due to the arbitrary, unexplained divergences stemming from \ul{the underlying \textbf{uncertainty} of workload or resource usage patterns}. Such uncertainties come from variability in the users' behavior, the impact of other users, uncertainty in resource availability, or even the sampling granularity of measurement.

Most existing research focus on specific scenarios instead of a principled, generalized formulation. For example, 
the ones on 
cloud oversubscription often try to solve the resource allocation problem via online bin-packing
~\cite{householder2014cloud} or leverage user migration for overload mitigation~\cite{wang2018energy,li2019adaptive}. 
Reinforcement learning with constraints~\cite{garcia2015comprehensive} can reasonably capture the dynamic usage patterns and uncertainty, 
but may not converge in a competing multi-objective (COGS vs risk) setting~\cite{achiam2017constrained,paternain2019constrained,mazyavkina2021reinforcement} and are hard to train/deploy on real systems. 

We employ Imitation Learning (IL), an offline policy learning strategy, to learn oversubscription policies leveraging historical resource usage telemetry data. 
But, such telemetry has significant aleatoric uncertainty which may cause divergence or high variance making policies unsafe against resource congestion risk. 
While Uncertainty Quantification (UQ) methods \cite{abdar2021review} can explicitly represent uncertainty in prediction or decision-making problems, they are not tractable for real-time adaptive oversubscription. They also cannot capture the real-world interactions between users and the service in production~\cite{de2019causal}. Also, \textbf{balance} between safety and gains is critical in real scenarios. 
Thus, to alleviate the above challenges,  we propose \ourmethod, a \textbf{\ul{C}}hance-c\textbf{\ul{\textsc{o}}}nstrained \textbf{\ul{\textsc{i}}}mitation lear\textbf{\ul{\textsc{n}}}ing framework that exploits  chance-constraints (stochastic) to implicitly model the underlying uncertainty and learn adaptive policies that not only optimize resource efficiency and reliability but are also safe and robust against congestion risk. 
IL with hard constraints, as we observe in the evaluations, might have strictly enforced safety, but will lead to substantially higher resource wastage. 
We make the following major contributions --
    \textbf{(1.)} We propose a novel chance-constrained imitation learning approach which, to the best of our knowledge, is a first of its kind. 
    \textbf{(2.)} We leverage the same for uncertainty-aware adaptive oversubscription to learn policies that are safe/robust against congestion risk while optimizing resource efficiency.
    \textbf{(3.)} We demonstrate, through extensive evaluations on oversubscription scenarios (Cloud services - $1^{st}$ party/internal and $3^{rd}$ party/public Azure, Airline ticketing, etc.), how \ourmethod~can learn safe yet effective policies. 



\section{Related Work}
\emph{Oversubscription:}
Cloud oversubscription is practiced for different resource types, \eg, CPU~\cite{cohen2019overcommitment}, memory~\cite{predClusterOversub}, power~\cite{kumbhare2021prediction}, etc. However, our evaluations focus on on virtual CPU/core oversubscription as  CPU is one of the most vital, albeit scarce, resources in the cloud with respect to VM allocation~\cite{protean}. Further details on related literature of industrially practiced solutions in oversubscription on cloud and other domains are presented in Appendix C.

\emph{Chance-Constrained RL:}
Chance constrained RL has gained popularity in recent years.
Some model-based chance-constrained RL methods improve the over-conservative policy with efficient evaluations~\cite{peng2021model,peng2022model}. 
However model-based methods are challenging to deploy on real systems. Other model-free chance-constrained RL methods primarily adopt penalty and Lagrangian techniques \cite{geibel2005risk,paternain2019learning}. 
However, the oversubscription problem involves a large number of chance constraints for sets of users, making existing methods difficult to adapt.

\emph{Safe Imitation Learning:}
Most traditional imitation learning, inverse RL or even Constrained IL approaches~\cite{ILwCons,diehl2022differentiable} struggle to induce safe policies in presence of uncertainty, which is even more critical in context of adaptive resource management. Some of the existing works that does study safe imitation learning from uncertain or inconsistent demonstrations focus on explicitly predicting/estimating uncertainty via kernels or model-based formulation \cite{SilverioEtAl2019,EnglertEtAl2013} and enforce safety by computing "value at risk". Some try to estimate via Monte Carlo policy rollouts or via ensembles (with dropouts) \cite{MendaEtAl2019,CuiEtAl2019}. While our approach may have partial similarity to the above, our stochastic constraints implicitly ensure safety against uncertainty instead of explicit estimation of the same and we also bypass the need to sample or rollout trajectories, thereby making it sample efficient. (Discussion on traditional IL in Appendix C)


\section{Method}

\subsection{Problem Setting}
\label{section:problemsetting}

\subsubsection{Uncertain Telemetry.}
Telemetry data in our setting comprises the resource usage measurements across time characterized by the relevant features about the system load. 
Note that there are no direct ground truth labels about ideal oversubscription margin/rate. Instead, we use the resource usage rate at current time as the surrogate ground truth label. The decision problem is to estimate a oversubscription rate (as a ratio of requested resources,  $\mathfrak{R}\in[0,100] = \%\text{ of requested resource}$) closely aligned with the resource usage rate, the surrogate ground truth.

\begin{figure}[t]
  \centering
  \includegraphics[width=0.95\linewidth]{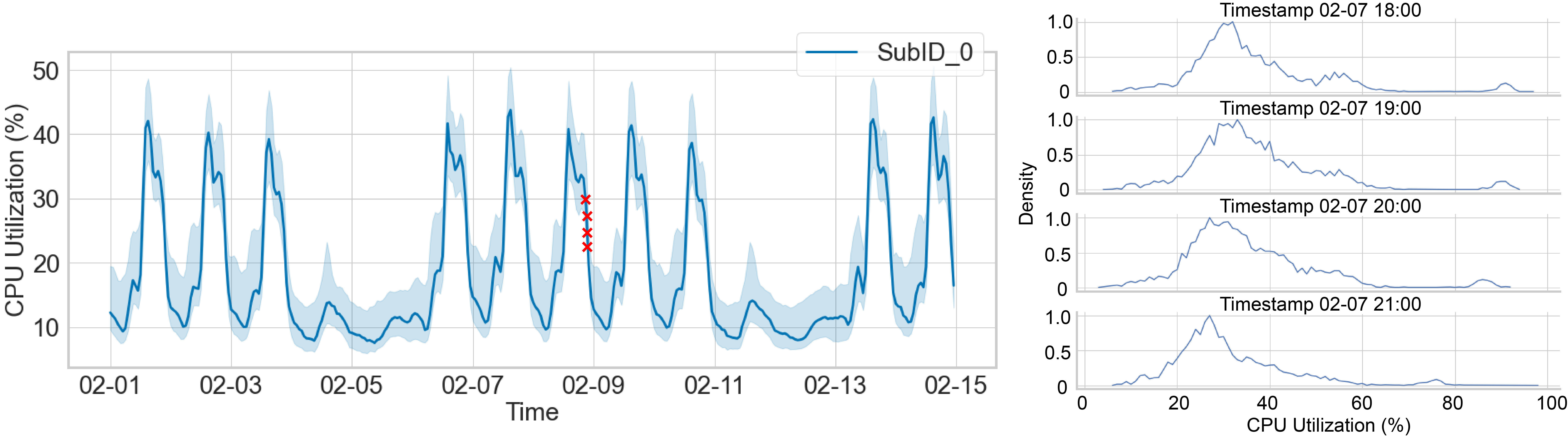}
  \caption{CPU utilization of a sampled subscriber (left), where the variance band represents $25^{th}$ and $75^{th}$ percentiles. Four points (marked as red crosses) are sampled to give their probability density function on the right. }
  \label{fig:Gaussian}
  \vspace{-4mm}
\end{figure}

The aleatoric uncertainties (AU) \cite{hullermeier2021aleatoric} present in such telemetry data stem from behavioral uncertainty of the resource usage of each user/entity, statistical uncertainties due to the granularity of measurement, and noise induced by aggregation at various granularities. Figure~\ref{fig:Gaussian} illustrates an example from cloud services domain, where the CPU usage rate for different subscribers/users at different time steps are not point estimates but are, approximately, characterized by Gaussian distributions (possibly skewed/mixture). 
These stochastic (non-point) estimates contribute to the AU. Resource scaling for reactive mitigation and the dependence among users' patterns add to the AU as well. 
Epistemic (model) uncertainty (EU) implicitly exists in any model that has not converged or has not observed sufficient samples. However, AU significantly worsens EU, even post convergence. 
This is even more critical in sequential decision-making scenario because  the cumulative unexplained variance across time will significantly increase resource congestion risk. 


\subsubsection{Imitation Learning from uncertain trajectories.}
Traditional imitation learning can be viewed as sequential supervised learning where the goal is to minimize the divergence between the agent policy $\pi_\theta$(predicted actions) and the expert's policy $\pi^E$ (underlying policy induced by the usage telemetry) [$L = \pi^E(a_t|s_t) \ominus \pi_\theta(\hat{a}_t|s_t)$], where $s_t \in S, a_t\in A$, $S$ and $A$ being the state space and action space respectively. 
However, as hinted earlier, when usage trajectories/data contain uncertainty, it becomes a harder problem. 
Considering the aleatory uncertainty in the data as a variance in the expert policy distribution, $\sigma_E^2 = \mathbb{E}_{s\sim T(S)}\left[\left((a\sim \pi^E(a|s))-\mathbb{E}_{a\sim \pi^E}[a|s]\right)^2\right]$, 
the bias-variance decomposition of the generic loss, temporarily assuming a squared error form, gives,
\begin{align}
    \nonumber L & = \pi^E(a_t|s_t) \ominus \pi_\theta(\hat{a}_t|s_t)\\
    \label{eqn:irredicible} & = \underbrace{(\mathbb{E}_{\hat{a}\sim\pi_\theta}[\hat{a}|s] - \mathbb{E}_{a\sim\pi^E}[a|s])^2}_{\text{bias}} + \underbrace{\sigma_\theta^2}_{\text{model var}}
   + \underbrace{(\sigma^E)^2}_{\text{irreducible}}
\end{align}
The aleatoric uncertainty shows up as the irreducible variance component in the error and in case of KL divergence loss the irreducible noise reduces to entropy (the information theoretic measure of uncertainty). \textit{This leads to either uncontrolled catastrophic resource congestion in out-of-distribution scenarios \cite{diehl2022differentiable} if we strictly follow the learned policy, or will reduce potential benefits if we add some safety margin.} It is difficult to directly mitigate irreducible noise/variance, especially in sequential decision-making problems. While some RL attempt to address this via constraints on action space to control exploration in feasible regions, others focus on explicit UQ \cite{Lockwood_Si_2022} via Bayesian RL. 
We do not need to measure or compute quantities like "value-at-risk" unlike some existing methods to enforce safety. 
\ourmethod~uses constraints with `probably approximate' satisfiability, (chance) on congestion risk metric (not on the action space) and presents a way to solve them tractably without trajectory sampling. 



\subsection{Problem Formulation and Solution Overview}
We propose the \textbf{\ul{C}}hance-c\textbf{\ul{\textsc{o}}}nstrained \textbf{\ul{\textsc{i}}}mitation lear\textbf{\ul{\textsc{n}}}ing (\ourmethod), where the expert's trajectory $\tau$ consisting of states and actions $(s_0,a_0,s_1,a_1,...,s_T, a_T)$ are drawn from the expert policy $\pi_E$. In oversubscription, we consider the user's resource requests, the historical resources usage, the current resource status, etc., as the states, and the current usage rate as the actions $a_t\in[0,1]$. We try to learn a policy $\pi_\theta(\hat{a}|s)$, where $\hat{a}$ is the predicted oversubscription rate. 
 Instead of learning a policy with soft constraints from in-distribution data (like vanilla IL), we aim to find a policy that mimics the expert policy while satisfying the chance constraints.  We formulate our problem as:
\begin{align}
\begin{split}
 \min_\pi ~& L(\pi^E, \pi_{\theta})\\
\text { s.t. } ~ & \operatorname{Pr}\left(\frac{1}{T}\sum\nolimits_{t=0}^T c(s_{t}) \leq g \right) \geq 1-\delta\\
 ~&  s_0\sim  \mathcal{P}_0,  s_{t+1} \sim \mathcal{P}(s_t, a_t);\mkern3mu a_{t} \sim \pi_\theta(\cdot|s_t); 
 \end{split}
 \label{eq: imcc}
\end{align}
where $L(\pi^E, \pi_{\theta})$ is the objective function of mimicking the expert's policy, which can be a mean square error function for Behavior Cloning \cite{pomerleau1991efficient} or a JS divergence loss function for adversarial imitation learning. In this paper, we adopt behavior cloning as the base learner for simplicity of implementation on real-world data. $c(s_t)=h^{\top}s_t$ is the instantaneous constraint cost which is predefined by the task. $\mathcal{P}_0$ is the distribution of $s_0$, $\mathcal{P}(s'|s, a)$ is the state transition probability for $s \rightarrow s'$ against action $a$. We consider a linear dynamic system $\mathcal{P}(s'|s, a)$ under Gaussian distribution for converting the chance constraints into deterministic form. Such Gaussian assumption is valid and follows directly from the usage rate being a Gaussian distribution as shown in Section~\ref{section:problemsetting} and allows for transforming into deterministic constraints. 
$T$ is the time horizon of the oversubscription task.
Equation~\ref{eq: imcc} is the chance constraint which indicates that the probability of satisfying cumulative constraints shall be more than $1-\delta$, where $\delta$ is the global upper bound of the probability of resource congestion. 
$g$ is the upper bound on the cumulative constraint cost, i.e., resource limit 
for the whole oversubscription action. 

\ul{Challenges:} The main challenge of learning $\pi_\theta$ is the constraint is stochastic and cumulative and depends on the distribution of entire trajectories', making the optimization problem much harder since -- \textbf{(I.)} Verifying satisfiability will require computing integral of a multi-variate probability distribution which are not only intractable to compute but the solution may also be non-existent if $c(s_t)\leq g_t$ is not always mutually exclusive across all states. \textbf{(II.)} Even measuring the probability values requires sampling of trajectories at every iteration, which compromises sample efficiency making it infeasible to be deployed in real systems. 



\begin{figure}[t!]
    \centering
    \includegraphics[width=\columnwidth]{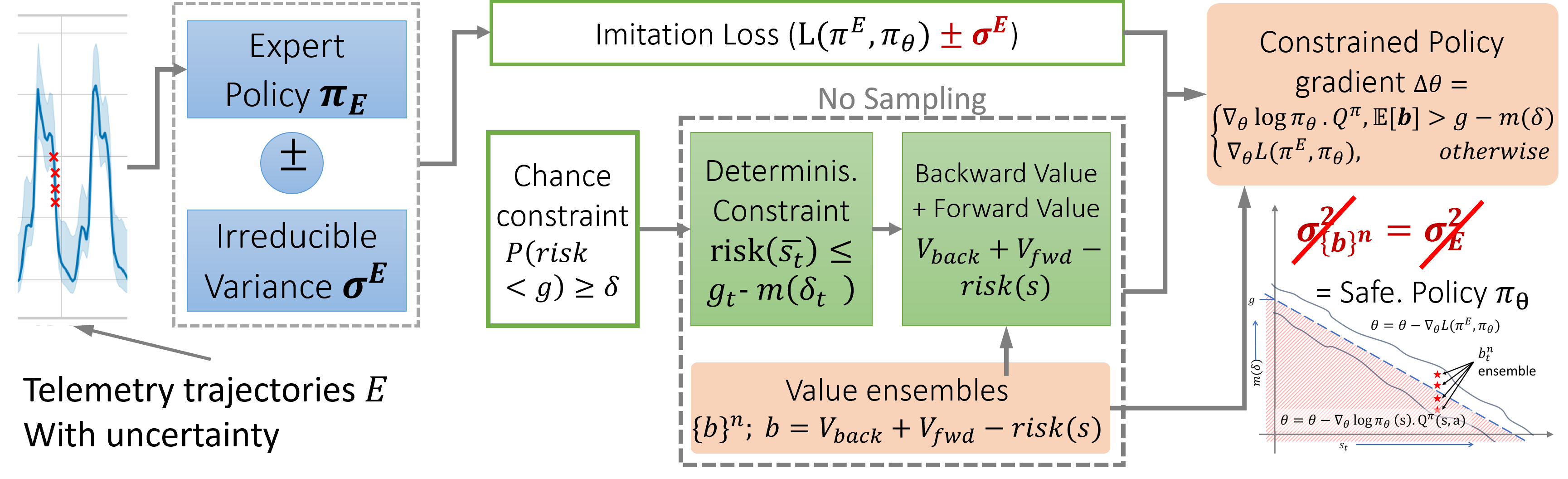}
    \caption{Solution Overview of \ourmethod}
    \label{fig:overview}
    \vspace{-1.5em}
\end{figure}

\ul{Solution Strategies:} We address these challenge via three mechanisms. 
\textit{(1) Stochastic $\rightarrow$ Deterministic [Challenge I]}: Due to the difficulty of computing the integral, we transform the chance constraint in Equation~\ref{eq: imcc} into a deterministic constraint defined on the states $\bar{s}_t$, where $\bar{s}_t = E[s_t]$ is the expected value of state (represents the major value of the states).
\textit{(2) Backward Value [Challenge II]}: Given the deterministic constraint, inspired by the previous work, we estimate the constraint value for each time step with a temporal difference (TD)~\cite{sutton2018reinforcement} methods instead of sampling entire trajectories. This method updates the estimated value of chance constraint via bootstrapping. 
\textit{(3) Value ensembles [Challenge II]}: We further design a practical tractable implementation of our proposed chance-constrained IL for real-world problems via value function ensembles. Figure~\ref{fig:overview} illustrates the solution schematic.

\subsection{Satisfying the chance constraint}
Unlike traditional constrained optimization, typical branch and bound strategies will not suffice in optimizing an IL objective. So we describe how to transform a chance constraint into a deterministic constraint to assert (stochastic) constraint satisfaction and construct the backward value function in the context of IL with the deterministic form. 

\subsubsection{Transforming to deterministic constraint}
With the assumption that the state transaction distribution is Gaussian distribution, i.e.,
$h_t^{\top}s_t \sim \mathcal{N}(h_t^{\top}\bar{s}_t,\sigma_t^2)$,
we can derive the deterministic version of the chance constraint as follows:
\begin{align}
    \label{eq: determisnisticConstraint}
    c(\bar{s}_t)\leq g_t - m(\delta_t)
\end{align}
where $\bar{s}_t = E[s_t]$ is the nominal state of the agent, $c(\bar{s}_t) = h_t^{\top}\bar{s}_t$, $m(\cdot)$ is the inverse of the cumulative distribution function of univariate Gaussian distribution, $g_t$ and $\delta_t$ are local cost and chance constraint probability per time step $t$. 
\begin{align}
\label{eqn:mdelta}
     m(\delta_t)  = - \sqrt{\sigma_t ^2}\Phi^{-1}(\delta_t),
\end{align}
where $\sigma_t = \sqrt{2h_t^{\top}\sum_{s_t}h_t}$ 
is the standard derivation and $\Phi^{-1}(\delta_t) = \texttt{erf}^{-1}(\delta_t )$ is the inverse of the Gaussian  function. 

\begin{lemma}[Deterministic $\approx$ Chance constraint]
\label{lemma:deter=chance}
A feasible solution to the deterministic constraint in Equation~\ref{eq: determisnisticConstraint} is always a feasible solution to the original chance constraint in Equation~\ref{eq: imcc}. (Proof in Appendix~A.2) 
\end{lemma}

\subsubsection{Backward value function for satisfiability estimation.}

In (chance-constrained) imitation learning, the policy needs to consider the cumulative constraints from time step $0$ to current time step $t$ to plan for the future.
Now, given the derived deterministic constraint function in Equation~\ref{eq: determisnisticConstraint}, we observe that the constraint is estimated by sampling the whole trajectory, which is not sample efficient. To improve the sample efficiency, TD learning estimates the current time step's cumulative value via bootstrapping the previous time step's cumulative value, i.e., $V^{\pi}(s_t)\leftarrow r(s_t,a_t)+\gamma \sum_{s_{t=1}}V^{\pi}(s_{t+1})$, where $V^{\pi}(s)$ is the state value function. This method is largely used in reinforcement learning to estimate the cumulative reward from the current time step to $T$. This cumulative constraint value is named as backward value function~\cite{satija2020constrained}, which estimates the at each time step via bootstrapping following the TD learning, which is extremely sampling efficient. Inspired by the previous work on backward value function, we design a new backward value function for the chance constraints. With the estimated constraint value, it can help to obtain an analytical solution step by step.



\begin{lemma}[Backward value $\thicksim$ Forward Markov chain]
With the irreducible and aperiodic assumption in forward Markov chain $\mathcal{P}^\pi(s_{t+1}|s_t) = \sum_{a_t\in \mathcal{A}}\mathcal{P}(s_{t+1}|s_t) \pi_\theta(a_t|s_t)$, samples from the forward Markov chain $\mathcal{P}^\pi(s_{t+1}|s_t)$ can be used directly to estimate the
statistics of the backward Markov chain $\overleftarrow{\mathcal{P}}^\pi(s_{t}|s_{t+1}) = \sum_{a_t\in \mathcal{A}}\overleftarrow{\mathcal{P}}(s_t,a_t|s_{t+1})$. For any $K \in \mathbf{N}$, where $\mathbf{N}$ is the set of natural numbers. Thus we can obtain the backward value function:
\begin{align}
   \overleftarrow{V}^\pi(s_t) & =  \mathbf{E}_{\overleftarrow{\mathcal{P}}^\pi}[\sum_{k=0}^Kc(s_{t-k})|s_t]\\
      & = \mathbf{E}_{\mathcal{P}^\pi,s_{t-K\sim \eta^\pi(\cdot) }}[\sum_{k=0}^Kc(s_{t-k})|s_t]
\end{align}
where $\mathbf{E}_{\overleftarrow{\mathcal{P}}}[\cdot]$ indicates the expectation over backwards chain, $\eta^\pi$ is the stationary distribution satisfying $\eta^\pi(s_{t+1}) = \sum_{s\in\mathcal{S}}\mathcal{P}^\pi(s_{t+1}|s_t)\eta^\pi(s_t)$. 
\end{lemma}

Based on lemma 2, we can get the backward value from the forward Markov chain instead of the backward Markov chain, which further simplifies the backward value estimation. We can then use the traditional TD learning setting to estimate the backward value function recursively as follows:
\begin{align}
 \overleftarrow{V}^\pi(s_t) = \mathbf{E}_{\overleftarrow{\mathcal{P}}^\pi}[c(s_t) + \overleftarrow{V}^\pi(s_{t-1})]
\end{align}

With the defined backward value function above, we can rewrite the derived deterministic constraint at each time step. Recall the derived deterministic constraint:
\begin{align}
\sum_{t=0}^Tc(\bar{s}_t)\leq g - m(\delta) \label{eq: original}
\rightarrow\mathbf{E}_{\mathcal{P}^\pi}[\sum_{t=0}^Tc(s_t)]\leq g - m(\delta)
\end{align}
Alternatively, for each time step $t\in[0,T]$ of the trajectory:
\begin{align}
\nonumber \underbrace{\mathbf{E}[\sum_{k=0}^tc(s_k)|s_0,\pi]}_{\text{backward from t}} + \underbrace{\mathbf{E}[\sum_{k=t}^Tc(s_k)]}_{\text{forward from t}}- \mathbf{E}[c(s_t)] \leq g - m(\delta)
\end{align}
The trajectory-level constraints can now be transformed into state-level,
\begin{align}
\mathbf{E}_{s_t\sim\eta^\pi(\cdot)}[\overleftarrow{V}^\pi(s_t) + 
V^\pi(s_t) - c(s_t)] \leq g - m(\delta).
\label{bvfconstraint}
\end{align}
As proved by the previous work, the state-level constraint in Eq.~\ref{bvfconstraint} with backward value function is an upper bound to the original constraint in Eq.\ref{eq: original}. Thus the policy satisfying the state-level constraint will also ensure the original constraint.

Given the formulation of the state-level constraint, we can obtain a policy improvement at each time step, where a safe feasible policy can be optimized by the estimated constraint value. The final objective with the state-level chance constraint imitation learning is as follows:
\begin{align}
\begin{split}
&\min_\pi ~ L(\pi^E, \pi_{\theta})\\
&\text{s.t.} ~ \mathbf{E}_{s_t\sim\eta^\pi}
[\overleftarrow{V}^\pi(s_t) + 
V^\pi(s_t) - c(s_t)] \leq g_t - m(\delta_t) 
\end{split}
\label{finalversion}
\end{align}


\subsection{Practical Implementation Design}
There are still two main challenges in optimizing the deterministic objective function of chance-constraint imitation learning in Eq.\ref{finalversion}. The first is that directly solving this constraint problem via Lagrangian-based methods or Lyapunov-based methods will introduce large extra computational complexity. The second is that when we try to check whether the policy is feasible in each time step, we need to compare the constraint value with the threshold which requires calculating the standard derivation. However, it is hard to estimate the standard derivation in large-scale datasets.

\subsubsection{Chance constrained optimization via policy gradient}
To solve the first challenge, we leverage the safety layers based method~\cite{dalal2018safe}, a constraint projection approach, to conduct action correction at each time step.
For each state, the main idea of the safety layer is to project the unconstrained action to the nearest action (in Euclidean norm) satisfying the necessary constraints. Following the derived analytical solution of previous work~\cite{satija2020constrained}, let $d(s) = \nabla Q^\pi(s,a )|_{a = \pi_\theta(s)}$, where $Q^\pi(s,a )$ is the state-action value: 
\begin{align}
    a* = \pi_\theta(s)-\lambda^*\cdot d(s), \label{projected_action}
\end{align}
where $\lambda^* = (\frac{-(g+c(s)-\overleftarrow{V}^\pi(s)-Q(s,\pi_\theta(s)))}{d(s)^{T}d(s)})^+$ is the derived analytical solution.

Based on the derived analytical solution, we can consider the constraint term as an additional loss, then conduct gradient descent to optimize the policy while the policy is not in feasible space, i.e., $\mathbf{E}_{s_t\sim\eta^\pi(\cdot)}[\overleftarrow{V}^\pi(s_t) + V^\pi(s_t) - c(s_t)] > g - m(\delta)$.
\begin{align}
    \theta \leftarrow \theta - \nabla_\theta\log \pi_\theta(s)\cdot Q^\pi(s,a).
\end{align}
Otherwise, we update the $\theta$ via the loss of traditional imitation learning:
\begin{align}
    \theta \leftarrow \theta - \nabla_\theta L(\pi^E,\pi_\theta).
\end{align}

\subsubsection{Ensemble Learning to estimate the variance of cost value}
To solve the second challenge, where we need to verify whether the policy is feasible in the current time step. We learn $N$ backward value function and $N$ forward value functions where they will sync with other during several steps of training to estimate the standard deviation of the costs.
\begin{align}
\label{eq:sd-ensemble}
    \sigma_t = \sum_{n=1}^N b^n_t - \mathbb{E}[b^n_t],
\end{align}
where $ b_t = \overleftarrow{V}^\pi(s_t) + 
V^\pi(s_t) - c(s_t)$ is the value of the state-level constraint. 

\begin{lemma}[Stochasticity $\models$ aleatoric uncertainty]
    The irreducible variance due to aleatoric uncertainty $\sigma^E$ (Eqn.~\ref{eqn:irredicible}) is  entailed by the model variance from the ensemble of value functions $\sigma_t$  in Eq.~\ref{eq:sd-ensemble}, $\sigma_t^2 \models (\sigma^E)^2$ where $\sigma_t^2 \propto K. m(\delta_t)^2$ [from Eqn.~\ref{eqn:mdelta}]. (Proof in Appendix A).
\end{lemma}

\subsection{Algorithm}
We integrate \ourmethod~on top of behavior cloning\footnote{Code files shared in supplementary material}. Details of the algorithm can be found in Algorithm~\ref{alg:algorithm1}. We discuss an alternate update for implementing the algorithms.

\begin{algorithm}[h!] 
	\caption{\ourmethod~(Safety Layer)}
        \begin{footnotesize}
	\begin{algorithmic}[1]
	 	\REQUIRE  \# epochs $M$,  $\pi_\theta$, $\overleftarrow{V_\omega}^\pi(s_t)$, $Q_\phi (s_t, a_t)$, constraint threshold $g$, constraint probability $\delta$;
		\STATE Initialize actor $\pi_\theta$, $N$ target backward value function $\overleftarrow{V_\omega}^\pi(s_t)^{n}$, target forward value function $Q_\phi (s_t, a_t)^n$, constraint threshold $g$ and upper bound of resource congestion probability $\delta$; 
		\FOR{$m=0$ to $M$} 
		\STATE Select the action $a_t$ from the the projection of action $a^*$ which is calculated by  Eq.\ref{projected_action}, take action $a_t$ and observe the state $s_{t+1}$ and cost $c(s_t)$;
          \STATE n $\leftarrow$ n+1;
		\STATE Calculate the backward value for the constraint with $N$ value functions;
  \STATE $Q^n(s_t, a_t)$, $V^n(s_t)$, $\overleftarrow{V_\omega}^{\pi,n}(s_t)$;
  \STATE Calculate the standard derivation of the value $\sigma$ and $m(\delta)$;
 \STATE Calculate the deterministic constraint to check whether the current action is feasible:
 \IF{$\overleftarrow{V}^\pi(s_t) + 
V^\pi(s_t) - c(s_t)] \leq g_t - m(\delta_t)$}
\STATE update the policy $\theta\leftarrow\theta-\nabla_\theta L(\pi^E,\pi_\theta) $
\ELSE
\STATE Use safeguard policy update to recover
\STATE $\theta\leftarrow\theta-\nabla_\theta\log\pi_\theta(s_t)Q(s_t)$
\ENDIF
\ENDFOR 
\STATE Update backward and forward constraint values here:
\FOR{$i\in\{t_{start},...,t\}$}
\STATE $\overleftarrow{R}\leftarrow c(s_t)+\gamma \overleftarrow{R}$
		\STATE $\overleftarrow{V}^{\pi,n}_\omega\leftarrow\phi-(\overleftarrow{R}-\overleftarrow{V}_\omega^{\pi,n})^2$ \hfill $\vartriangleright$ upd. backward const. val.
  \ENDFOR 
  \FOR{$i\in\{t-1,..,t_{start}\}$}
  \STATE $R\leftarrow c(s_t)+\gamma R$
  \STATE $Q^{n}_\phi\leftarrow Q^n_\phi-(R-Q^n_\phi)^2$ \hfill $\vartriangleright$ upd. forward constraint value
  \ENDFOR
	\end{algorithmic}
 \label{alg:algorithm1}
 \end{footnotesize}
\end{algorithm}

\section{Experiments}
We evaluate our proposed approach, \ourmethod, based on the following questions: 
\begin{enumerate}
    \item Does \ourmethod~outperform baselines in learning suitable oversubscription policies that lead to more efficient utilization in real systems/services? 
    \item How well does \ourmethod~satisfy the constraints on resource congestion risk to ensure safe policy?
    \item What is the post-convergence stability of \ourmethod~in presence of uncertainty as opposed to vanilla IL? 
\end{enumerate}

\textbf{Baselines: } In this paper, we compare our approach against the following baselines (1) Grid search with different oversubscription probabilities, where all the subscribers have the same oversubscription rate (2) Vanilla IL such as Behavior Cloning or BC, (3) Policy gradient reinforcement learning such as DDPG~\cite{lillicrap2015continuous} (4) Multi-Agent reinforcement learning or MA~\cite{sheng2022learning} and (5) IL with hard constraints. Most existing resource management or oversubscription frameworks in practice \cite{7553036,gosavi2002airline,kumbhare2021prediction,shihab2019autonomous,LAWHEAD2019252} exploit traditional reinforcement learning as the optimization strategy. Thus our baseline choices are reasonably comprehensive.

\textbf{Metrics:}
(1) \textbf{Cloud services:} The chance-safety of cloud service is evaluated via \textit{physical Machine hot ratio} (short as \textbf{PM-Hot-R}). In addition, we set chance-safety ratio $g$ (g $g \in \{\textbf{0.75, 0.85, 0.95}\})$ to check whether the oversubscription policy satisfies the chance constraint with different $g$.
PM-Hot-R indicates the maximum probability of a PM in the cluster being hot than $\delta$. 
The remaining cores quantity is evaluated via \textit{saved cores} (short as \textbf{S-Cores}).
(2) \textbf{Airline ticket overbooking:} The chance-safety of Airline ticket overbooking is evaluated via \textit{Ticket-Cost ratio} (short as \textbf{Ticket-Cost-R}). In addition, we set chance-safety ratio $g$ ( $g \in \{\textbf{0.75, 0.85, 0.95}\})$ to check whether the oversubscription policy satisfies the chance constraint with different $g$.
The profit is evaluated via \textit{Profit}.


\subsection{Experiment Design}
\subsubsection{Evaluation Domains} We evaluate \ourmethod~on 2 major domains, vCPU oversubscription in cloud services and flight tickets overbooking in Airlines. 

\textbf{Cloud Oversubscription:} 
In this domain we employ two different data sets: Z company's internal cloud\footnote{Orgnization name hidden for anonymity.} platform and a public dataset from Azure~\cite{azure}. Usage patterns in these two datasets differ, allowing us to evaluate the performance of our method under different conditions. Internal Cloud dataset contains sampled traces in February 2022, consisting of 1.5 million virtual machines (VMs). The AzurePublicDatasetV2 contains data from the 2019 Azure VM workload, comprising information on approximately 2.6 million VMs and 1.9 billion utilization readings. 

\textbf{Airline ticket overbooking:} In this scenario, flight overbooking, \ie, selling more tickets than the available seats, is a common practice that allows airline companies to improve their load factors and increase revenues~\cite{nazifi2021proactive}. Yet, the difficulty in estimating ticket demands and no-shows results in inappropriate overbooking strategies, such that users with tickets cannot onboard, \ie, offloaded. In general, flight ticket demands show quarter patterns that peak tourist seasons have higher flight demands~\cite{suryani2010air,banerjee2020passenger}. This motivates adaptive oversubscription of flight tickets.
We collect airline passengers' data from the overbooking reports of the U.S. Department of Transportation (DOT). The dataset covers overbooking information of 32 airline companies in the U.S. from 1998 to 2021\footnote{https://www.bts.gov/denied-confirmed-space}, reported quarterly. Each quarter's data includes an offloaded number of passengers (voluntary/involuntary) and the onboard number of passengers. 
Evaluating a decision-making framework like ours requires not just test datasets, but also an environment where we can observe feedback from actions. To that end we designed Oversub Gym environment. 

\textbf{Generality study:} Additional evaluations of \ourmethod~against Safe RL/IL baselines on Mujoco benchmarks is presented in Appendix A, to highlight that \ourmethod~can effectively generalize to any domain.

\subsubsection{Environment Settings}


Oversub Gym for cloud includes two environments built on two different cloud datasets.
We conduct experiments under both cold-start and warm-start scenarios in the Internal Cloud dataset. The cold-start scenario simulates an empty environment where no previous VMs have been allocated, while the warm-start scenario simulates an environment with previously allocated VMs running. It is important to note that we only investigate cold- and warm-start scenarios in the Internal Cloud dataset, as we were unable to obtain detailed information on previously allocated VMs from AzurePublicDatasetV2\footnote{\url{https://github.com/Azure/AzurePublicDataset}}. Our Oversub Gym environment allows for both the above schemes. The design of the environment involves the following critical aspects, namely, \textbf{(1) Allocation} of VMs to PMs - closely approximates real allocators used in production 
\textbf{(2) Reset}, for in environment intialization with warm/cold start contexts, 
\textbf{(3) State}, current status of the environment, \textbf{(4) Transition model} to represent state transitions, \textbf{(5) Action}, which represents the oversubscription rates and \textbf{(6)The chance constraints}, characterized by the cumulative constraints $g$, (e.g., in cloud service, $g$ indicates the ratio of \textbf{hot} physical machines (w/ high cpu usage rate)), and $\delta$ the probability of chance constraint satisfaction (Further details in Appendix B). For Airline ticket overbooking, we design a simulator trained with GBDT on the
semi-synthetic data (described earlier) that outputs the overbooking rate of the next quarter given current
quarter features.

\subsection{Experimental Results}
We present and discuss or experimental results with respect to each of the domains and data sets and we highlight the observations and insights based on the evaluation questions we listed earlier. 
\subsubsection{Internal Cloud}
We train our model with $1800$ episodes and $10$ seeds. 
Three different safety-levels, i.e., $g$ are considered as $0.75$, $0.85$, and $0.95$ respectively. 
Based on the given results of seven models on the internal cloud system Azure with the chance constraint probability $1-\delta = 0.95$, the following observations can be made:
(1) \ourmethod~satisfies chance constraint with all different chance-safety $g$ and achieves the highest saving than the other methods that also satisfy the constraints, i.e., Grid-0.6, MA, IL and IL with hard constraints. These results indicate that the chance constraint is a crucial factor in ensuring performance and safety together.  (2) \ourmethod~shows significantly more remain cores that IL+hard constraint, which indicates hard constraints are too conservative. (3) DDPG can MA shows worse performance in remain cores, which indicates that reinforcement learning would introduce more challenge in optimizing. 
 
These results highlight how \ourmethod~is extremely effective in designing profitable oversubscription policies in internal cloud system where COGS benefit is very important, answering \textbf{Q1} affirmatively while stochastically ensuring safety against congestion (hot node) risk (\textbf{Q1}). 

\subsubsection{Azure Public Cloud}
The average usage rate of Azure is much higher than the internal Cloud. 
As shown in Table~\ref{tb: azure-tb}, we can see that Grid fails to satisfy the safe constraint due to that most of the users have large usage rates.
Four methods, i.e., \ourmethod, IL, IL+hard constraint and MA,  satisfy the chance constraint with different chance-safety $g$. Our method \ourmethod~consistently achieves the largest remain cores among these four methods. The imitation learning based models achieve better performance than reinforcement learning based models, the reason would be that optimizing reinforcement learning objectives in the highly stochastic environment, i.e., cloud service, is hard.

In conclusion, the results also suggests that the chance constraints can help improve the performance ofimitation learning models.  These results highlight how \ourmethod is extremely effective in designing profitable oversubscription policies in external cloud system where COGS benefit is very important, answering \textbf{Q1} affirmatively. They also show that our approach can seamlessly satisfy the congestion constraints, affirming \textbf{Q2}.

\begin{table}[h]
    \centering
    \caption{Results in Internal Cloud}
    \resizebox{0.95\columnwidth}{!}{
    \begin{tabular}{cccccc}
    \toprule
    Method         & PM-Hot-R & S-Cores &  $0.75$ & $0.85$ & $0.95$   \\ \midrule
    Grid-0.2       & $95.4$                                 & $80.0$          & \redx & \redx & \redx        \\ 
    Grid-0.4       & $100$                                    & $60.0$          & \redx & \redx & \redx        \\ 
    Grid-0.6       & $0.0$                                      & $40.0$         & \greencheck & \greencheck & \greencheck         \\ 
    MA             & $0.0$                                     & $38.4$        & \greencheck & \greencheck & \greencheck        \\ 
    DDPG             & $6.7$                                     & $35.6\pm1.2$        & \greencheck & \greencheck & \greencheck        \\ 
    IL (BC)             & $0.8\pm1.1$                            & $49.9\pm3.3$              & \greencheck & \greencheck & \greencheck        \\ 
    IL + hard constraint             & $0.73\pm1.5$                            & $37.9\pm4.6$              & \greencheck & \greencheck & \greencheck        \\ 
    \ourmethod & $1.5\pm 0.97$               & $66.8\pm 13.5$    & \greencheck & \greencheck & \greencheck  \\ 
    \bottomrule
    \end{tabular}}
    \label{tb: internal_tb}
\end{table}

\begin{table}[h]
    \centering
    \caption{Results in Azure public Cloud}
    \resizebox{0.95\columnwidth}{!}{
    \begin{tabular}{cccccc}
    \toprule
    Method         & PM-Hot-R & S-Cores &  $0.75$ & $0.85$ & $0.95$   \\ 
    \midrule
    Grid-0.2       & $100.0$                                 & $80.0$          & \redx & \redx & \redx        \\ 
    Grid-0.4       & $100.0$                               & $60.0$ & \redx &\redx & \redx        \\ 
    Grid-0.6       & $100.0$                                  & $40.0$         & \redx & \redx & \redx         \\ 
    MA                       & $0.0$                & $5.7$         & \greencheck & \greencheck  & \greencheck    \\ 
        DDPG                       & $3.89$                & $8.4$         & \greencheck & \greencheck  & \greencheck    \\ 
    IL (BC)            & $1.0\pm0.0$                                        & $7.6\pm2.2$        & \greencheck & \greencheck & \greencheck        \\ 
    IL+hard constraint             & $0.0\pm0.0$                                        & $5.4\pm2.1$        & \greencheck & \greencheck & \greencheck        \\ 
    \ourmethod & $1.67\pm 3.2$    & $15.78\pm0.13$    & \greencheck & \greencheck & \greencheck  
    \\
    \bottomrule
    \end{tabular}}
    \label{tb: azure-tb}
\end{table}

\subsubsection{Airline Tickets Overbooking.}
As the overbooking rate of each airline company and the actual demands are not reported, we generate overbooking rate and flight demands based on known common industry practice~\cite{suzuki2006net} to create a semi-synthetic dataset. 
With the popularity of electronic tickets, no-shows are less compared to the paper-ticket period~\cite{nazifi2021proactive}, we assume the no-show rate to decay with years. Then the demand is the aggregation of bumped, no-shows, and onboard passengers. Our GBDT-based simulator is trained on the above semi-synthetic data that outputs the overbooking rate of the next quarter given current quarter features. 
The results are shown in Table~\ref{tb:airline-tb}. We observe that, (1) the Grid methods fail, where none of them satisfy the chance constraints, which indicates that global oversubscription rates are not suitable for this task. (2) \ourmethod~shows siginificant improvements in profits compared to baselines while satisfying the chance constraints on Ticket-Cost ratio. These results highlight how \ourmethod~is extremely effective in designing profitable oversubscription policies in Ticket booking task answering \textbf{Q1} affirmatively. Thus, our approach satisfies the stochastic constraints congestion constraints ensuring balanced safety, affirming \textbf{Q2}. 
\begin{table}[t]
\centering
\caption{Oversubscription results in Airline}
\resizebox{0.95\columnwidth}{!}{
\begin{tabular}{cccccc}
\toprule
Method         & Ticket-Cost-R & Profit &  $0.75$ & $0.85$ & $0.95$   \\ 
\midrule
Grid-0.2       & $100.0$                                 & $16.24$          & \redx & \redx & \redx        \\ 

Grid-0.4       & $100.0$                               & $12.18$ & \redx &\redx & \redx        \\ 
Grid-0.6       & $100.0$                                  & $8.12$         & \redx & \redx & \redx         \\ 
MA                       & $1.7\%$                & $7.2$         & \greencheck & \greencheck  & \greencheck    \\ 
DDPG                       & $1.28\%$                & $6.9$         & \greencheck & \greencheck  & \greencheck    \\ 
IL (BC)            & $0.9\%\pm0.0$                                        & $7.8\pm2.2$        & \greencheck & \greencheck & \greencheck        \\ 
IL+hard constraint             & $0.0\pm0.0$                                        & $5.4\pm2.1$        & \greencheck & \greencheck & \greencheck        \\ 
\ourmethod & $0.3\%\pm 3.2$    & $9.78\pm0.13$    & \greencheck & \greencheck & \greencheck  

\\
\bottomrule
\end{tabular}
}
\label{tb:airline-tb}
\end{table}

\begin{figure}[t]
    \centering
    \subfigure[Average Mean Square Error]{
    \includegraphics[width=0.43\columnwidth]{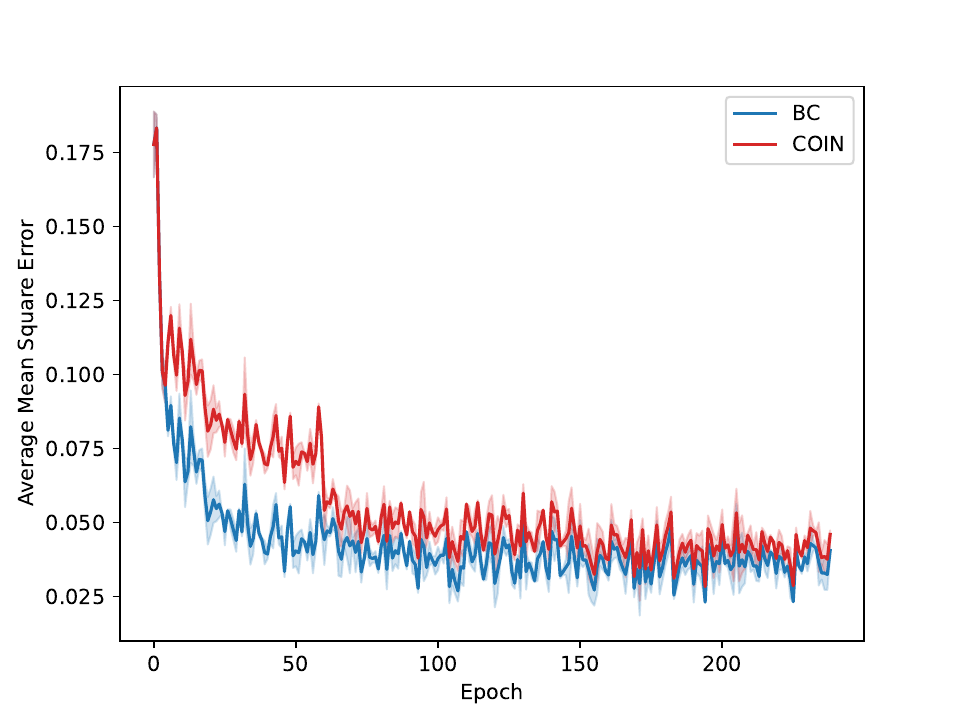}
    \label{fig:mseCloud}
    }
    \subfigure[Average Hot nodes]{
    \includegraphics[width=0.43\columnwidth]{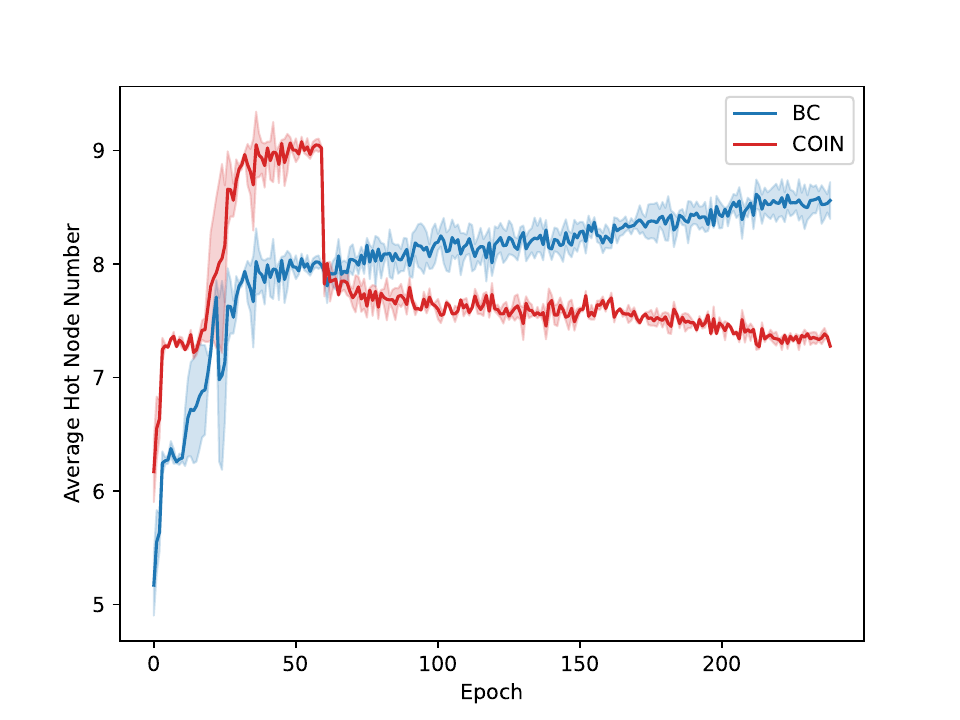}
    \label{fig:nm_hot_cluster}
    }
    \caption{Convergence curves for vCPU Oversubscription}
    \label{fig:conv_curves_cloud}
\end{figure}

\begin{figure}[t]
    \centering
    \subfigure[Mean Square Error.]{
    \includegraphics[width=0.43\columnwidth]{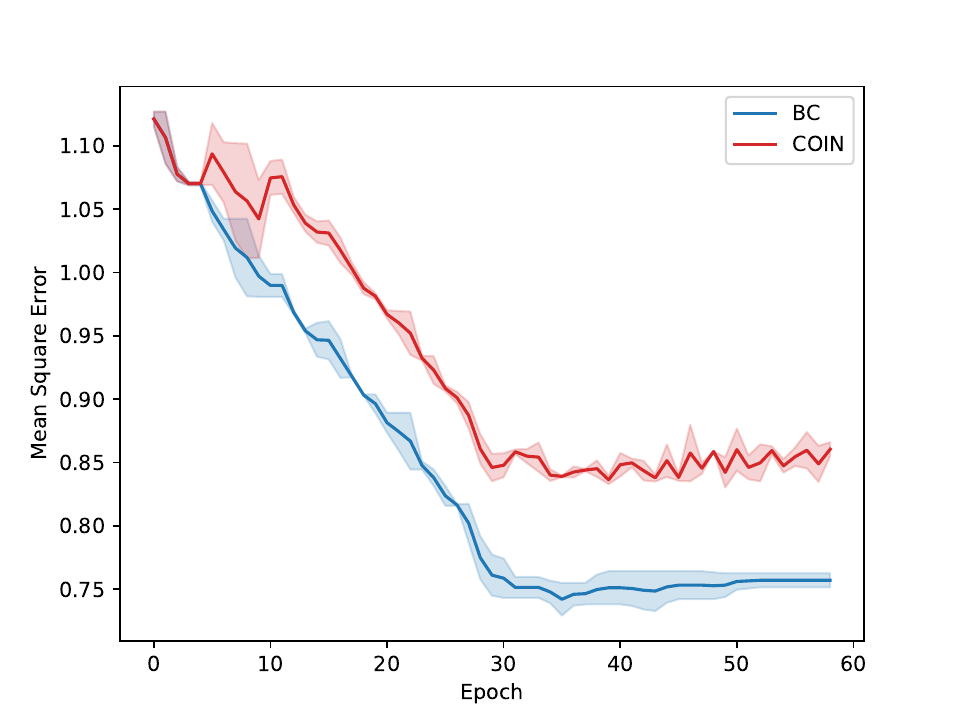}
    \label{fig:msefly}
    }
    \subfigure[Average Cost]{
    \includegraphics[width=0.43\columnwidth]{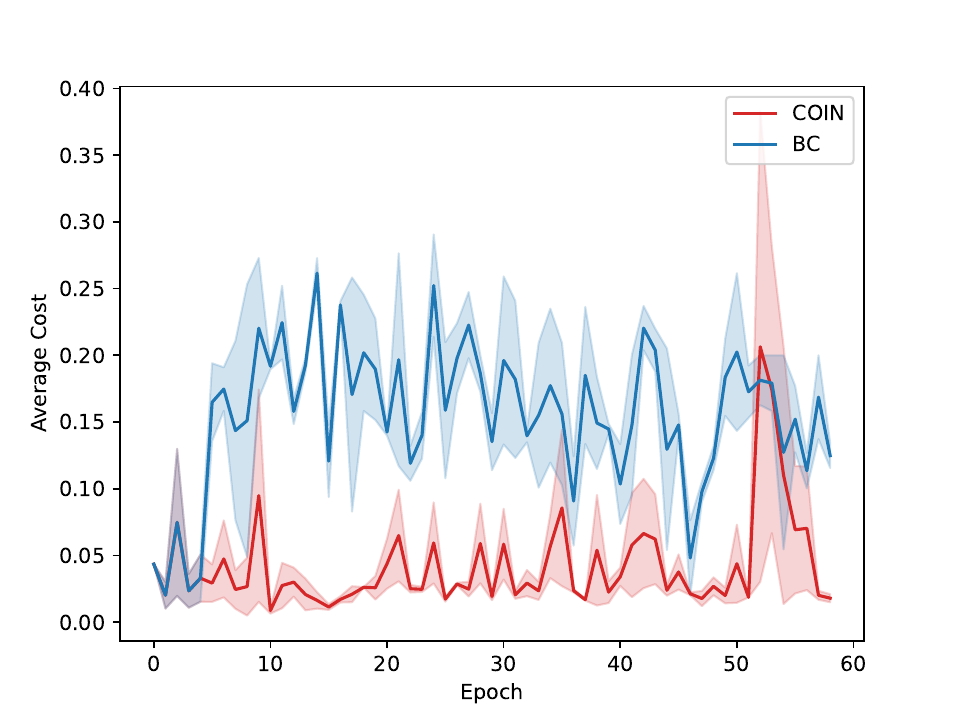}
    \label{fig:Costfly}
    }
    \caption{Convergence curves for Airline Overbooking}
    \label{fig:conv_curves_air}
\end{figure}

\subsection{Convergence Analysis}

Adaptive uncertainty-aware oversubscription framework should lead to better operational saving, without compromising on real service reliability post-deployment. To that end our policy learner should have stable convergence properties, such that (1) the overhead cost of training policies does not exceed the savings in operational cost and (2) The policy is well balanced and ensures both "safety" and saving. Convergence analysis on both vCPU oversubscription as well as Air ticket overbooking domains under uncertainty and compare \ourmethod 
 against a stable baseline (Behavior Cloning or BC) is presented in Figure~\ref{fig:mseCloud} \&~\ref{fig:nm_hot_cluster}.

\ourmethod~converges to a stable mean squared error almost at the same epoch range as the baseline BC in the Cloud vCPU oversubscription scenario. Note however, that BC drops faster towards the earlier epochs. This indicates that training with stochastic constraints may be a bit slower and early-stopping could lead to sub-optimal results. We observe similar behavior in the average hot node number curve (Fig~\ref{fig:nm_hot_cluster}). 
However, \ourmethod~achieves significantly better (lower) average hot nodes at convergence. This shows that it can appropriately but stochastically satisfy the chance constraint on hot nodes ensuring balanced safety. 
Figure~\ref{fig:Costfly} shows that (1) BC continuously has the higher average costs than \ourmethod~ on airline data. 
Thus, \ourmethod~has beneficial convergence properties under uncertainty unlike vanilla imitation learning baseline, affirmatively answering \textbf{Q3}.




\section{Conclusion}
We presented our `first of its kind' chance-constrained imitation learning framework that can implicitly model uncertainty and learn balanced safe yes profitable oversubscription policies. That leads to substantial COGS benefit/profit with nominal congestion risk. Adaptive oversubscription with stochastic bounds on risk is very important in real systems since industry cannot adopt strict risk avoidance (hard constraints) yet make reasonable profits. 
As future work we plan to design an end-to-end oversubscription framework that can not only leverage utilization patterns but can also exploit workload forecasts. Furthermore we will show the effectiveness of our proposed approach on other domains beyond the ones highlighted in this paper. 

\bibliography{kr-sample}

\begin{thebibliography}{43}
\providecommand{\natexlab}[1]{#1}

\bibitem[{Abdar et~al.(2021)Abdar, Pourpanah, Hussain, Rezazadegan, Liu, Ghavamzadeh, Fieguth, Cao, Khosravi, Acharya et~al.}]{abdar2021review}
Abdar, M.; Pourpanah, F.; Hussain, S.; Rezazadegan, D.; Liu, L.; Ghavamzadeh, M.; Fieguth, P.; Cao, X.; Khosravi, A.; Acharya, U.~R.; et~al. 2021.
\newblock A review of uncertainty quantification in deep learning: Techniques, applications and challenges.
\newblock \emph{Information Fusion}, 76: 243--297.

\bibitem[{Achiam et~al.(2017)Achiam, Held, Tamar, and Abbeel}]{achiam2017constrained}
Achiam, J.; Held, D.; Tamar, A.; and Abbeel, P. 2017.
\newblock Constrained policy optimization.
\newblock In \emph{International conference on machine learning}, 22--31. PMLR.

\bibitem[{Banerjee, Morton, and Akartunal{\i}(2020)}]{banerjee2020passenger}
Banerjee, N.; Morton, A.; and Akartunal{\i}, K. 2020.
\newblock Passenger demand forecasting in scheduled transportation.
\newblock \emph{European Journal of Operational Research}, 286(3): 797--810.

\bibitem[{Baset, Wang, and Tang(2012)}]{baset2012towards}
Baset, S.~A.; Wang, L.; and Tang, C. 2012.
\newblock Towards an understanding of oversubscription in cloud.
\newblock In \emph{2nd USENIX Workshop on Hot Topics in Management of Internet, Cloud, and Enterprise Networks and Services (Hot-ICE 12)}.

\bibitem[{Breitgand and Epstein(2012)}]{breitgand2012improving}
Breitgand, D.; and Epstein, A. 2012.
\newblock Improving consolidation of virtual machines with risk-aware bandwidth oversubscription in compute clouds.
\newblock In \emph{2012 Proceedings IEEE INFOCOM}, 2861--2865. IEEE.

\bibitem[{Chen et~al.(2018)Chen, Cao, Zhang, Ma, Zhou, and Yang}]{predClusterOversub}
Chen, J.; Cao, C.; Zhang, Y.; Ma, X.; Zhou, H.; and Yang, C. 2018.
\newblock Improving cluster resource efficiency with oversubscription.
\newblock In \emph{2018 IEEE 42nd Annual Computer Software and Applications Conference (COMPSAC)}, volume~1, 144--153. IEEE.

\bibitem[{Cohen et~al.(2019)Cohen, Keller, Mirrokni, and Zadimoghaddam}]{cohen2019overcommitment}
Cohen, M.~C.; Keller, P.~W.; Mirrokni, V.; and Zadimoghaddam, M. 2019.
\newblock Overcommitment in cloud services: Bin packing with chance constraints.
\newblock \emph{Management Science}, 65(7): 3255--3271.

\bibitem[{Cui et~al.(2019)Cui, Isele, Niekum, and Fujimura}]{CuiEtAl2019}
Cui, Y.; Isele, D.; Niekum, S.; and Fujimura, K. 2019.
\newblock Uncertainty-Aware Data Aggregation for Deep Imitation Learning.
\newblock In \emph{2019 International Conference on Robotics and Automation (ICRA)}, 761--767.

\bibitem[{Dalal et~al.(2018)Dalal, Dvijotham, Vecerik, Hester, Paduraru, and Tassa}]{dalal2018safe}
Dalal, G.; Dvijotham, K.; Vecerik, M.; Hester, T.; Paduraru, C.; and Tassa, Y. 2018.
\newblock Safe exploration in continuous action spaces.
\newblock \emph{arXiv preprint arXiv:1801.08757}.

\bibitem[{De~Haan, Jayaraman, and Levine(2019)}]{de2019causal}
De~Haan, P.; Jayaraman, D.; and Levine, S. 2019.
\newblock Causal confusion in imitation learning.
\newblock \emph{Advances in Neural Information Processing Systems}, 32.

\bibitem[{Diehl et~al.(2022)Diehl, Adamek, Kr{\"u}ger, Hoffmann, and Bertram}]{diehl2022differentiable}
Diehl, C.; Adamek, J.; Kr{\"u}ger, M.; Hoffmann, F.; and Bertram, T. 2022.
\newblock Differentiable Constrained Imitation Learning for Robot Motion Planning and Control.
\newblock \emph{arXiv preprint arXiv:2210.11796}.

\bibitem[{Englert et~al.(2013)Englert, Paraschos, Peters, and Deisenroth}]{EnglertEtAl2013}
Englert, P.; Paraschos, A.; Peters, J.; and Deisenroth, M.~P. 2013.
\newblock Model-based imitation learning by probabilistic trajectory matching.
\newblock In \emph{2013 IEEE International Conference on Robotics and Automation}, 1922--1927.

\bibitem[{Garc{\i}a and Fern{\'a}ndez(2015)}]{garcia2015comprehensive}
Garc{\i}a, J.; and Fern{\'a}ndez, F. 2015.
\newblock A comprehensive survey on safe reinforcement learning.
\newblock \emph{Journal of Machine Learning Research}, 16(1): 1437--1480.

\bibitem[{Geibel and Wysotzki(2005)}]{geibel2005risk}
Geibel, P.; and Wysotzki, F. 2005.
\newblock Risk-sensitive reinforcement learning applied to control under constraints.
\newblock \emph{Journal of Artificial Intelligence Research}, 24: 81--108.

\bibitem[{Gosavi, Bandla, and Das(2002)}]{gosavi2002airline}
Gosavi, A.; Bandla, N.; and Das, T. 2002.
\newblock Airline seat allocation among multiple fare classes with overbooking.
\newblock \emph{IIE Transactions}, 34(9): 729--742.

\bibitem[{Hadary et~al.(2020)Hadary, Marshall, Menache, Pan, Greeff, Dion, Dorminey, Joshi, Chen, Russinovich, and Moscibroda}]{protean}
Hadary, O.; Marshall, L.; Menache, I.; Pan, A.; Greeff, E.~E.; Dion, D.; Dorminey, S.; Joshi, S.; Chen, Y.; Russinovich, M.; and Moscibroda, T. 2020.
\newblock Protean: {VM} Allocation Service at Scale.
\newblock In \emph{14th USENIX Symposium on Operating Systems Design and Implementation (OSDI 20)}, 845--861. USENIX Association.
\newblock ISBN 978-1-939133-19-9.

\bibitem[{Householder, Arnold, and Green(2014)}]{householder2014cloud}
Householder, R.; Arnold, S.; and Green, R. 2014.
\newblock On cloud-based oversubscription.
\newblock \emph{arXiv preprint arXiv:1402.4758}.

\bibitem[{H{\"u}llermeier and Waegeman(2021)}]{hullermeier2021aleatoric}
H{\"u}llermeier, E.; and Waegeman, W. 2021.
\newblock Aleatoric and epistemic uncertainty in machine learning: An introduction to concepts and methods.
\newblock \emph{Machine Learning}, 110: 457--506.

\bibitem[{Kawaharazuka et~al.(2021)Kawaharazuka, Kawamura, Okada, and Inaba}]{ILwCons}
Kawaharazuka, K.; Kawamura, Y.; Okada, K.; and Inaba, M. 2021.
\newblock Imitation Learning With Additional Constraints on Motion Style Using Parametric Bias.
\newblock \emph{IEEE Robotics and Automation Letters}, 6(3): 5897--5904.

\bibitem[{Kumbhare et~al.(2021)Kumbhare, Azimi, Manousakis, Bonde, Frujeri, Mahalingam, Misra, Javadi, Schroeder, Fontoura et~al.}]{kumbhare2021prediction}
Kumbhare, A.~G.; Azimi, R.; Manousakis, I.; Bonde, A.; Frujeri, F.; Mahalingam, N.; Misra, P.~A.; Javadi, S.~A.; Schroeder, B.; Fontoura, M.; et~al. 2021.
\newblock $\{$Prediction-Based$\}$ Power Oversubscription in Cloud Platforms.
\newblock In \emph{2021 USENIX Annual Technical Conference (USENIX ATC 21)}, 473--487.

\bibitem[{Lawhead and Gosavi(2019)}]{LAWHEAD2019252}
Lawhead, R.~J.; and Gosavi, A. 2019.
\newblock A bounded actor–critic reinforcement learning algorithm applied to airline revenue management.
\newblock \emph{Engineering Applications of Artificial Intelligence}, 82: 252--262.

\bibitem[{Li(2019)}]{li2019adaptive}
Li, Z. 2019.
\newblock An adaptive overload threshold selection process using Markov decision processes of virtual machine in cloud data center.
\newblock \emph{Cluster Computing}, 22(2): 3821--3833.

\bibitem[{Lillicrap et~al.(2015)Lillicrap, Hunt, Pritzel, Heess, Erez, Tassa, Silver, and Wierstra}]{lillicrap2015continuous}
Lillicrap, T.~P.; Hunt, J.~J.; Pritzel, A.; Heess, N.; Erez, T.; Tassa, Y.; Silver, D.; and Wierstra, D. 2015.
\newblock Continuous control with deep reinforcement learning.
\newblock \emph{arXiv preprint arXiv:1509.02971}.

\bibitem[{Lockwood and Si(2022)}]{Lockwood_Si_2022}
Lockwood, O.; and Si, M. 2022.
\newblock A Review of Uncertainty for Deep Reinforcement Learning.
\newblock In \emph{Proceedings of the AAAI Conference on Artificial Intelligence and Interactive Digital Entertainment}.

\bibitem[{Mazyavkina et~al.(2021)Mazyavkina, Sviridov, Ivanov, and Burnaev}]{mazyavkina2021reinforcement}
Mazyavkina, N.; Sviridov, S.; Ivanov, S.; and Burnaev, E. 2021.
\newblock Reinforcement learning for combinatorial optimization: A survey.
\newblock \emph{Computers \& Operations Research}, 134: 105400.

\bibitem[{Menda, Driggs-Campbell, and Kochenderfer(2019)}]{MendaEtAl2019}
Menda, K.; Driggs-Campbell, K.; and Kochenderfer, M.~J. 2019.
\newblock EnsembleDAgger: A Bayesian Approach to Safe Imitation Learning.
\newblock In \emph{2019 IEEE/RSJ International Conference on Intelligent Robots and Systems (IROS)}, 5041--5048.

\bibitem[{Nazifi et~al.(2021)Nazifi, Gelbrich, Gr{\'e}goire, Koch, El-Manstrly, and Wirtz}]{nazifi2021proactive}
Nazifi, A.; Gelbrich, K.; Gr{\'e}goire, Y.; Koch, S.; El-Manstrly, D.; and Wirtz, J. 2021.
\newblock Proactive handling of flight overbooking: how to reduce negative eWOM and the costs of bumping customers.
\newblock \emph{Journal of Service Research}, 24(2): 206--225.

\bibitem[{Paternain et~al.(2019{\natexlab{a}})Paternain, Calvo-Fullana, Chamon, and Ribeiro}]{paternain2019learning}
Paternain, S.; Calvo-Fullana, M.; Chamon, L.~F.; and Ribeiro, A. 2019{\natexlab{a}}.
\newblock Learning safe policies via primal-dual methods.
\newblock In \emph{2019 IEEE 58th Conference on Decision and Control (CDC)}, 6491--6497. IEEE.

\bibitem[{Paternain et~al.(2019{\natexlab{b}})Paternain, Chamon, Calvo-Fullana, and Ribeiro}]{paternain2019constrained}
Paternain, S.; Chamon, L.; Calvo-Fullana, M.; and Ribeiro, A. 2019{\natexlab{b}}.
\newblock Constrained reinforcement learning has zero duality gap.
\newblock \emph{Advances in Neural Information Processing Systems}, 32.

\bibitem[{Peng et~al.(2022)Peng, Duan, Chen, Li, Xie, Zhang, Guan, Mu, and Sun}]{peng2022model}
Peng, B.; Duan, J.; Chen, J.; Li, S.~E.; Xie, G.; Zhang, C.; Guan, Y.; Mu, Y.; and Sun, E. 2022.
\newblock Model-Based Chance-Constrained Reinforcement Learning via Separated Proportional-Integral Lagrangian.
\newblock \emph{IEEE Transactions on Neural Networks and Learning Systems}.

\bibitem[{Peng et~al.(2021)Peng, Mu, Guan, Li, Yin, and Chen}]{peng2021model}
Peng, B.; Mu, Y.; Guan, Y.; Li, S.~E.; Yin, Y.; and Chen, J. 2021.
\newblock Model-based actor-critic with chance constraint for stochastic system.
\newblock In \emph{2021 60th IEEE Conference on Decision and Control (CDC)}, 4694--4700. IEEE.

\bibitem[{Pomerleau(1991)}]{pomerleau1991efficient}
Pomerleau, D.~A. 1991.
\newblock Efficient training of artificial neural networks for autonomous navigation.
\newblock \emph{Neural computation}.

\bibitem[{Salahuddin, Al-Fuqaha, and Guizani(2016)}]{7553036}
Salahuddin, M.~A.; Al-Fuqaha, A.; and Guizani, M. 2016.
\newblock Reinforcement learning for resource provisioning in the vehicular cloud.
\newblock \emph{IEEE Wireless Communications}, 23(4): 128--135.

\bibitem[{Satija, Amortila, and Pineau(2020)}]{satija2020constrained}
Satija, H.; Amortila, P.; and Pineau, J. 2020.
\newblock Constrained markov decision processes via backward value functions.
\newblock In \emph{International Conference on Machine Learning}, 8502--8511. PMLR.

\bibitem[{Shahrad et~al.(2020)Shahrad, Fonseca, Goiri, Chaudhry, Batum, Cooke, Laureano, Tresness, Russinovich, and Bianchini}]{azure}
Shahrad, M.; Fonseca, R.; Goiri, I.; Chaudhry, G.; Batum, P.; Cooke, J.; Laureano, E.; Tresness, C.; Russinovich, M.; and Bianchini, R. 2020.
\newblock Serverless in the Wild: Characterizing and Optimizing the Serverless Workload at a Large Cloud Provider.
\newblock In \emph{2020 USENIX Annual Technical Conference (USENIX ATC 20)}, 205--218. USENIX Association.
\newblock ISBN 978-1-939133-14-4.

\bibitem[{Sheng et~al.(2022)Sheng, Wang, Yang, Qiao, Dong, Wang, Jin, Wang, Qin, Rajmohan et~al.}]{sheng2022learning}
Sheng, J.; Wang, L.; Yang, F.; Qiao, B.; Dong, H.; Wang, X.; Jin, B.; Wang, J.; Qin, S.; Rajmohan, S.; et~al. 2022.
\newblock Learning Cooperative Oversubscription for Cloud by Chance-Constrained Multi-Agent Reinforcement Learning.
\newblock \emph{arXiv preprint arXiv:2211.11759}.

\bibitem[{Shihab et~al.(2019)Shihab, Logemann, Thomas, and Wei}]{shihab2019autonomous}
Shihab, S. A.~M.; Logemann, C.; Thomas, D.-G.; and Wei, P. 2019.
\newblock Autonomous airline revenue management: A deep reinforcement learning approach to seat inventory control and overbooking.
\newblock \emph{arXiv preprint arXiv:1902.06824}.

\bibitem[{Shlifer and Vard(1975)}]{airlineoverbooking1975}
Shlifer, E.; and Vard, Y. 1975.
\newblock An Airline Overbooking Policy.
\newblock \emph{Transportation Science}, 9(2): 101--114.

\bibitem[{Silvério et~al.(2019)Silvério, Huang, Abu-Dakka, Rozo, and Caldwell}]{SilverioEtAl2019}
Silvério, J.; Huang, Y.; Abu-Dakka, F.~J.; Rozo, L.; and Caldwell, D.~G. 2019.
\newblock Uncertainty-Aware Imitation Learning using Kernelized Movement Primitives.
\newblock In \emph{2019 IEEE/RSJ International Conference on Intelligent Robots and Systems (IROS)}.

\bibitem[{Suryani, Chou, and Chen(2010)}]{suryani2010air}
Suryani, E.; Chou, S.-Y.; and Chen, C.-H. 2010.
\newblock Air passenger demand forecasting and passenger terminal capacity expansion: A system dynamics framework.
\newblock \emph{Expert Systems with Applications}, 37(3): 2324--2339.

\bibitem[{Sutton and Barto(2018)}]{sutton2018reinforcement}
Sutton, R.~S.; and Barto, A.~G. 2018.
\newblock \emph{Reinforcement learning: An introduction}.
\newblock MIT press.

\bibitem[{Suzuki(2006)}]{suzuki2006net}
Suzuki, Y. 2006.
\newblock The net benefit of airline overbooking.
\newblock \emph{Transportation Research Part E: Logistics and Transportation Review}, 42(1): 1--19.

\bibitem[{Wang and Tianfield(2018)}]{wang2018energy}
Wang, H.; and Tianfield, H. 2018.
\newblock Energy-aware dynamic virtual machine consolidation for cloud datacenters.
\newblock \emph{IEEE Access}, 6: 15259--15273.

\end{thebibliography}

\end{document}